%% file: ms.tex
\documentclass[11pt,a4paper]{article}
\usepackage[nohyperref]{conll2018}
\usepackage{times}
\usepackage{latexsym}

\usepackage{url}

\usepackage{graphicx}
\usepackage{amsmath,amsthm,amssymb,amsfonts}
\usepackage{caption}
\usepackage{subcaption}

\usepackage{array}
\usepackage{todonotes}
\newcolumntype{L}[1]{>{\raggedright\let\newline\\\arraybackslash\hspace{0pt}}m{#1}}
\newcolumntype{C}[1]{>{\centering\let\newline\\\arraybackslash\hspace{0pt}}m{#1}}
\newcolumntype{R}[1]{>{\raggedleft\let\newline\\\arraybackslash\hspace{0pt}}m{#1}}

\aclfinalcopy 

\title{\textit{Upcycle} Your OCR: Reusing OCRs for Post-OCR Text Correction in Romanised Sanskrit}

\author{Amrith Krishna$^{\#}$, Bodhisattwa Prasad Majumder\textsuperscript{*}, Rajesh Shreedhar Bhat\textsuperscript{**}, \\ \textbf{and Pawan Goyal}$^{\#}$ \\
  \textsuperscript{\#}Dept. of Computer Science and Engineering, IIT Kharagpur, \\\textsuperscript{*}Dept. of Computer Science, University of California, San Diego \\\textsuperscript{**}Walmart Labs, India \\
  \tt amrith@iitkgp.ac.in, bmajumde@eng.ucsd.edu,\\ \tt rajeshbhatpesit@gmail.com, pawang@cse.iitkgp.ernet.in}

\date{}

\begin{document}
\maketitle
\begin{abstract}
We propose a post-OCR text correction approach for digitising texts in Romanised Sanskrit. Owing to the lack of resources our approach uses OCR models trained for other languages written in Roman. Currently, there exists no dataset available for Romanised Sanskrit OCR. So, we bootstrap a dataset of 430 images, scanned in two different settings and their corresponding ground truth. For training, we synthetically generate training images for both the settings. We find that the use of copying mechanism \cite{gu-EtAl:2016:P16-1} yields a percentage increase of 7.69 in Character Recognition Rate (CRR) than the current state of the art model in solving monotone sequence-to-sequence tasks \cite{schnober2016still}.  We find that our system is robust in combating OCR-prone errors, as it obtains a CRR of 87.01\% from an OCR output with CRR of 35.76\% for one of the dataset settings. A human judgement survey performed on the models shows that our proposed model results in predictions which are faster to comprehend and faster to improve for a human than the other systems\footnote{The data and the codes for our system are available here - \url{https://github.com/majumderb/sanskrit-ocr}}.
\end{abstract}

\input{1intro.tex}
\input{1task.tex}
\input{2training.tex}

\input{1results.tex}

\bibliography{conll2018}
\bibliographystyle{acl_natbib_nourl}

\end{document}

%% file: 1intro.tex
\section{Introduction}

Sanskrit used to be the `lingua franca' for the scientific and philosophical discourse in ancient India with literature that spans more than 3 millennia. Sanskrit primarily had an oral tradition, and the script used for writing Sanskrit varied widely across the time spans and regions. With the advent of printing press, Devanagari emerged as the prominent script for representing Sanskrit. With standardisation of Romanisation using IAST in 1894 \cite{dictionary1899monier}, printing in Sanskrit was extended to roman scripts as well. There has been a surge in digitising printed Sanskrit manuscripts written in Roman such as the ones currently digitised by the `Krishna Path' project\footnote{\url{http://www.krishnapath.org/library/}}. 

\begin{figure}[h]
\vspace{-1em}
    \centering
    \includegraphics[width=\linewidth]{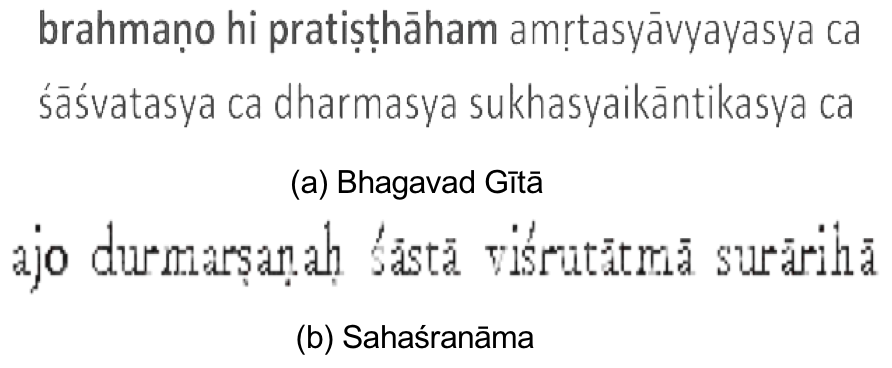}
    \caption{Sample images from our test set with different stylistic parameters}
    \label{fig:test}
    \vspace{-1em}
\end{figure}

In this work, we propose a model for post-OCR text correction for Sanskrit written in Roman. Post-OCR text correction, which can be seen as a special case of spelling correction \cite{schnober2016still}, is the task of correcting errors that tend to appear in the output of the OCR in the process of converting an image to text. The errors incurred from OCR can be quite high due to numerous factors including typefaces, paper quality, scan quality, etc. The text can often be eroded, can contain noises and the paper can be bleached or tainted as well~\cite{schnober2016still}. Figure~\ref{fig:test} shows the sample images we have collected for the task. Hence it is beneficial to perform a post-processing on the OCR output to obtain an improved text.

In the case of Indic OCRs, there have been considerable efforts in collection and annotation of data pertaining to Indic Scripts \cite{kumar2007content,bhaskarabhatla2004representation,govindaraju2009guide,krishnan2014towards}. Earlier attempts on Indian scripts were primarily based on handcrafted templates \cite{govindan1990character,chaudhuri1997ocr} or features~\cite{arora2010recognition,pal2009comparative} which extensively used the script and language-specific information \cite{krishnan2014towards}. Sequential labelling approaches were later proposed that take the word level inputs and make character level predictions \cite{shaw2008offline,hellwig2015ind}. The word based sequence labelling approaches were further extended to use neural architectures, especially using RNNs and its variants such as LSTMs and GRUs \cite{sankaran2012recognition, krishnan2014towards, saluja2017error, saluja2018wsc, mathew2016multilingual}. But, OCR is putative in exhibiting few long-range dependencies \cite{schnober2016still}.  \newcite{singh2015can} find that extending the neural models to process the text at the sentence level (or a textline) leads to improvement in the performance of the OCR systems. This was further corroborated by \newcite{saluja2017error} where the authors found that using words within a context window of 5 for a given input word worked particularly well for the Post-OCR text correction in Sanskrit. 
In the case of providing a text line as input, we are essentially providing more context about the input in comparison to the word level models and the RNN (or LSTM) cells are powerful enough to capture the long-term dependencies. Particularly for Indian languages, this decision is beyond a question of performance. 
In Sanskrit, the word boundaries are often obscured due to phonetic transformations at the word boundaries known as Sandhi. Word segmentation of Sanskrit constructions is a matter of research on its own~\cite{krishna-EtAl:2016:COLING,reddy2018building}. However, none of the existing systems are equipped for incorrect spellings and hence these systems may be brittle~\cite{belinkov2018synthetic} when it comes to handling spelling variations in the input. 
Hence, in our case, we assume an unsegmented sequence as our input and then we perform our Post-OCR text correction on the text. We hypothesise that this will improve the segmentation process and other downstream tasks for Sanskrit in a typical NLP pipeline.

\if{}
\begin{figure}[h]
    \centering
    \includegraphics[trim={0 0 0 0},clip,width=\linewidth]{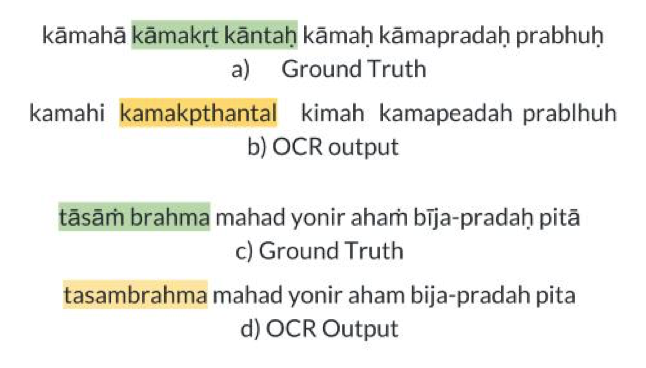}
    \caption{OCRs where word boundaries are detected improperly}
    \label{fig:sample}
    \vspace{-1em}
\end{figure}
\fi

Our major contributions are:

\begin{enumerate}

\item Contrary to what is observed in \newcite{schnober2016still}, an encoder-decoder model, when equipped with copying mechanism \cite{gu-EtAl:2016:P16-1}, can outperform a traditional sequence labelling model in a monotone sequence labelling task. Our model outperforms \newcite{schnober2016still} in the Post-OCR text correction for Romanised Sanskrit task by 7.69 \% in terms of CRR.

\item  By making use of digitised Sanskrit texts, we generate images as synthetic training data for our models. We systematically incorporate various distortions to those images so as to emulate the settings of the original images.

\item Through a human judgement experiment, we asked the participants to correct the mistakes from a predicted output from the competing systems. We find that participants were able to correct predictions from our system more frequently and the corrections were done much faster than the CRF model by \newcite{schnober2016still}. We observe that predictions from our model score high on acceptability \cite{lau-clark-lappin:2015:ACL-IJCNLP} than other methods as well. 



\end{enumerate}

%% file: 1task.tex
\section{Model Architecture}
\label{model}

In principle, the output from any OCR which recognises Romanised Sanskrit can be used as the input to our model. 
Currently, there exist limited options for recognising Romanised Sanskrit texts from scanned documents. Possibly, the commercial OCR offering by Google as part of their proprietary cloud vision API and SanskritOCR\footnote{\url{https://sri.auroville.org/projects/sanskrit-ocr/}. It provides interface to tesseract and Google OCR as well.} might be the only two viable options. 
SanskritOCR provides an online interface to the Tesseract OCR, an open source multilingual OCR \cite{smith2007overview,smith2009adapting,smith1987extraction}, trained specifically for recognising Romanised Sanskrit.  Additionally, we trained an offline version of Tesseract to recognise the graphemes in the Romanised Sanskrit alphabet. In both the models we find that many scanned images, especially similar to the one shown in Figure \ref{fig:test}b, were not recognised by the system. We hypothesise this to be due to lack of enough font styles available in our collection, in spite of using a site with the richest collection of Sanskrit fonts\footnote{More details about the training procedure in \S 1 of the supplementary material}. This leaves the Google OCR as the only option.

Considering the fact that working with a commercial offering from Google OCR may not be an affordable option for various digitisation projects, we chose to use Tesseract with models trained for other languages written in Roman script. All the Latin or Roman scripts in the pre-trained models of Tesseract are trained on 400,000 text-lines spanning about 4500 fonts\footnote{\url{https://github.com/tesseract-ocr/tesseract/wiki/TrainingTesseract-4.00}}. 

\begin{figure}[h!]
\small
    \centering
    \includegraphics[width=\columnwidth]{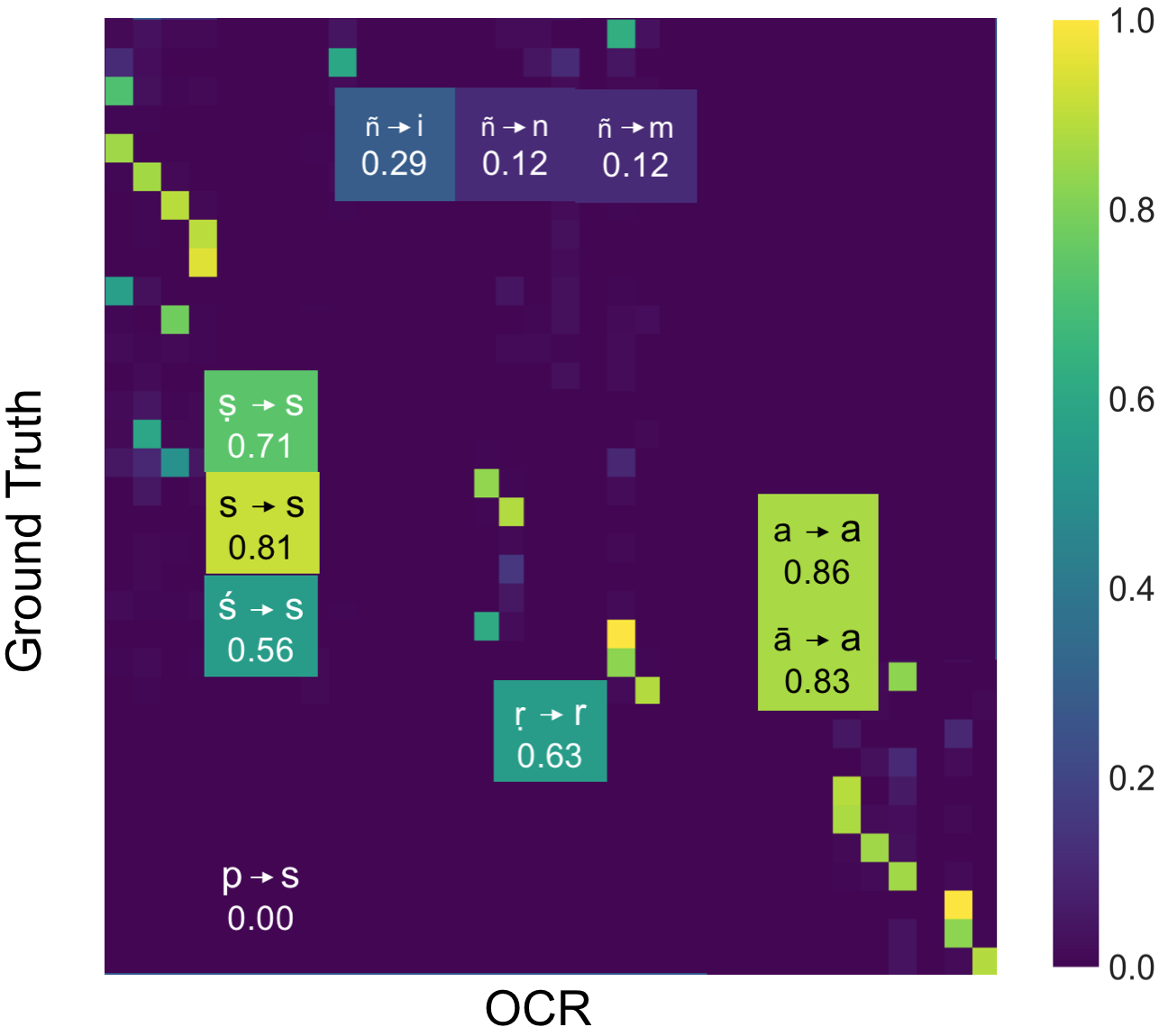}
    \caption{Heatmap of occurrences of majorly confusing character pairs between Ground Truth and OCR}
    \label{fig:ocr_errors}
    \vspace{-1em}
\end{figure}

\paragraph{Use of OCR with pre-trained models for other languages}
French alphabet has the highest grapheme overlap with that of the Sanskrit alphabet (37 of 50), while all other languages have one less grapheme common with Sanskrit. Hence, we arbitrarily take 5 of the languages in addition to French and perform our analysis. Table \ref{tab:lang} shows the character recognition rate (CRR) for OCR using alphabets of different languages, when performed on a dataset of 430 scanned images (\S \ref{dataset}). The table also shows the count of error types made by the OCR after alignment \cite{jiampojamarn2007:,dhondt-grouin-grau:2016:LOUHI}. All the languages have a near similar CRR with English and French leading the list. Based on our observations on the OCR performance, we select English for our further experiments.

Upcycling such a pre-trained model brings its own challenges. For instance, the missing 14 Sanskrit graphemes\footnote{Detailed in \S 2 of the Supplementary Material} in English are naturally mispredicted to other graphemes. This leads to ambiguity as the correct and the mispredicted characters now share the same target. Figure \ref{fig:ocr_errors} shows the heat-map for such mis-predictions when we used the OCR on the set of 430 scanned images. Here, we zoom the relevant cases and show the row-normalised proportion of predictions\footnote{A more detailed figure with all the cases are available in the supplementary material in \S 3.}.

\begin{table*}[t!]
\small
\centering
\begin{tabular}{|c|l|l|l|l|l|l|l|l|c|}
\hline
\textbf{Language}             & \multicolumn{4}{c|}{\textbf{Bhagavad G{\=i}t{\=a}}} & \multicolumn{4}{c|}{\textbf{Saha\'{s}ran{\=a}ma}} & \multicolumn{1}{c|}{\textbf{Combined}} \\ \hline
\multicolumn{1}{|l|}{}        &  CRR  & Ins  & Del  & Sub    &  CRR  & Ins& Del & Sub  &   CRR      \\ \hline
English                       & 84.92 & 23   & 63   & 1868   & 64.06 & 73 & 696 & 1596 &  80.08     \\ \hline
French                        & 84.90 & 21   & 102  & 1710   & 63.91 & 91 & 702 & 1670 &  80.04      \\ \hline
Finnish                       & 82.61 & 15   & 141  & 1902   & 61.31 & 80 & 730 & 1821 &  78.81        \\ \hline
Italian                       & 83.45 & 20   & 73   & 1821   & 62.19 & 84 & 690 & 1673 &  79.03     \\ \hline
Irish                         & 84.52 & 12   & 78   & 1810   & 63.81 & 72 & 709 & 1841 &  79.93        \\ \hline
German                        & 84.40 & 33   & 72   & 1821   & 63.79 & 87 & 723 & 1874 &  79.12     \\ \hline
\end{tabular}
\caption{OCR performances for different languages with overall CRR, total Insertion, Deletion and Substitution errors.}
\label{tab:lang}
\vspace{-1em}
\end{table*}

\subsection{System Descriptions}

We formalise the task as a monotone seq2seq model. We use an encoder-decoder framework that takes in a character sequence as input and the model finds embeddings at a sub-word level both at the encoder and decoder side. Here the OCR output forms input to the model. Keeping the task in mind we make two design decisions for the model. One is the use of copying mechanism \cite{gu-EtAl:2016:P16-1} and other is the use of Byte Pair Encoding (BPE) \cite{sennrich-haddow-birch:2016:P16-12} to learn a new vocabulary for the model.

\paragraph{CopyNet \cite{gu-EtAl:2016:P16-1}:} Since it is possible that there will be reasonable overlap between the input and output strings, we use the copying mechanism as mentioned in CopyNet \cite{gu-EtAl:2016:P16-1}. The model essentially learns two probability distributions, one for generating an entry at the decoder and the other for copying the entry from the encoder. The final prediction is based on the sum of both the probabilities for the class. Given an input sequence $\boldsymbol{X}=(\boldsymbol{x_1}, ..., \boldsymbol{x_{N}})$ we define $\mathcal{X}$, for all the \emph{unique} entries in the input sequence. We also define the vocabulary $\mathcal{V}=\{v_1, ..., v_N\}$. Let the out-of-vocabulary (OOV) words be represented with \textsc{unk}. The probability of the generate mode $g$ and copy mode $c$ are given by
\vspace{-1em}

\begin{eqnarray*} 
\label{eq:pg}
p(\boldsymbol{y_t}, \textsf{ g}|\cdot) 
\hspace{-3pt } 
=&
\hspace{-11pt} 
 \left \{\begin{matrix}
\dfrac{1}{Z}e^{\psi_g(\boldsymbol{y_t} )},                       \;\;\; \qquad& \;\;\; \boldsymbol{y_t}  \in \mathcal{V} \\
0,                                                                          \;\;\; \qquad&\;\;\; \boldsymbol{y_t} \in \mathcal{X} - \mathcal{V}\\
\dfrac{1}{Z}e^{\psi_g(\textsc{unk})} \;\;\;\;     \;\;\; \qquad& \;\;\; \boldsymbol{y_t}  \not \in \mathcal{V} \cup \mathcal{X}     \\
\end{matrix}\right. \\
\label{eq:pc}
	p(\boldsymbol{y_t}, \textsf{ c}|\cdot)
	\hspace{-3pt } 
	=& 
	\hspace{-12pt} 
	\left \{\begin{matrix}
\dfrac{1}{Z}\sum_{j:\boldsymbol{x_j}=\boldsymbol{y_t}} e^{\psi_c(\boldsymbol{x_j})},  \hspace{-17pt} & \boldsymbol{y_t} \in \mathcal{X} \\
0 &\text{otherwise} 
\end{matrix}\right. 
\end{eqnarray*}
where $\psi_g(\cdot)$ and $\psi_c(\cdot)$ are score functions for generate-mode ($g$) and copy-mode ($c$), respectively, and $Z$ is the normalization term shared by the two modes, $Z = \sum_{v\in \cal V \cup \{\textsc{unk}\}}e^{\psi_g(v)} + \sum_{\boldsymbol{x}\in \boldsymbol{X}}e^{\psi_c(\boldsymbol{x})}.$
The scoring function for both the modes, respectively, are
\vspace{-1em}

\begin{equation*}\label{eq:gen}
	\psi_g(\boldsymbol{y_t}=v_i) = \boldsymbol{v}_i^\top \boldsymbol{W_o} \boldsymbol{s_t}, \quad v_i \in \mathcal{V} \cup \textsc{unk}
\end{equation*}
\vspace{-1em}
\begin{equation*}\label{eq:cp}
	\psi_c(\boldsymbol{y_t}=\boldsymbol{x_j}) =	\sigma\left(\mathbf{h}_j^\top \mathbf{W}_c\right)\mathbf{s}_t, \quad  \boldsymbol{x_j} \in \mathcal{X} 
\end{equation*}

where $\mathbf{W}_c\in \mathbb{R}^{d_h \times d_s}$, and $\sigma$ is a non-linear activation function \cite{gu-EtAl:2016:P16-1}.

\paragraph{BPE \cite{sennrich-haddow-birch:2016:P16-12}}: Sanskrit is a morphologically rich language. A noun in Sanskrit can have 72 different inflections and a verb may have more than 90 inflections. Additionally, Sanskrit corpora generally express a compound rich vocabulary \cite{krishna-EtAl:2016:WSSANLP2016}. Hence, in a typical Sanskrit corpus, the majority of the tokens appear less than 5 times (\S \ref{dataset}). These are generally considered to be rare words in a corpus \cite{sennrich-haddow-birch:2016:P16-12}. However, corpora dominated by rare words are difficult to handle for a statistical model like ours. 
To combat the sparsity of the data, we convert the tokens into sub-word n-grams using Byte Pair Encoding (BPE) \cite{sennrich-haddow-birch:2016:P16-12}. Methods such as wordpiece \cite{Schuster6289079} as well as \newcite{sennrich-haddow-birch:2016:P16-12} are means of obtaining a new vocabulary for a given corpus. Every sequence in the corpus is then re-written as a sequence of tokens in terms of the sub-word units which forms the type in the new vocabulary so obtained. These methods essentially use a data-driven approach to maximise the language-model likelihood of the training data, given an evolving word definition \cite{wu2016google}.

We explicitly set the minimum count for a token in the new vocabulary to appear in the corpora as 30. We learn a new vocabulary of size 82 with 22 of them having a length 1 and the rest with a length 2. The IAST standardisation of the Romanised Sanskrit contains 50 graphemes in Sanskrit alphabet. About 12 of the graphemes are represented using 2 character roman character combinations. Now, in the vocabulary learnt using BPE, 7 of the graphemes were not present. Hence, we add them in addition to the 82 entries learnt as vocabulary. This makes the total vocabulary to be 89. By using the new vocabulary, it is guaranteed that there will be no Out Of Vocabulary (OOV) words in our model. 

We use 3 stacked layers of LSTM at the encoder and the decoder with the same settings as in \newcite{bahdanau2014neural}. To enable copying, we share the embeddings of the source and the target vocabulary. By eliminating OOV, we make sure that copying always happens by virtue of the evidence from the training data and not by the presence of an OOV word. 

%% file: 2training.tex
\section{Experiments}

\subsection{Dataset}
\label{dataset}

Sanskrit is a low-resource language. It is extremely scarce to obtain datasets with scanned images and the corresponding aligned texts for Romanised Sanskrit. We obtain 430 scanned images as shown in Figure \ref{fig:test} and manually annotate the corresponding text. We use this as our test dataset, henceforth to be referred to as \textit{OCRTest}.
For training, we synthetically generate images from digitised Sanskrit texts and use them as our training set and development set. The images for training, \textit{OCRTrain}, were generated by synthetically adding distortions to those images to match the settings of the real scanned documents.

\paragraph{OCRTest} contains 430 images from 1) scanned copy of Vishnu Saha\'{s}ran{\=a}ma\footnote{\url{http://kirtimukha.com/}} and 2) scanned copy of Bhagavad G{\=\i}t{\=a}, a sample of each is shown in Figure \ref{fig:test}a and \ref{fig:test}b. 
140 out of these 430 are from Saha\'{s}ran{\=a}ma and the remaining are from Bhagavad G{\=\i}t{\=a}. 

\paragraph{OCRTrain:} Similar to \newcite{Ul-Hasan:2013:WBL:2505377.2505394}, we synthetically generate the images, which are then fed to the OCR, to obtain our training data. We use the digitised text from \'{S}r{\=\i}mad Bh{\=a}gavatam\footnote{\url{https://www.vedabase.com/en/sb}} for generating the synthetic images. The text contains about 14,094 verses in total, divided into 50,971 text-lines. The dataset is divided into 80-20 split as training set and development set, respectively.  The corpus contains a vocabulary of 52,882 word types. 48,249 of the word types in the vocabulary appear less than or equal to 5 times, of which 32,411 appear exactly once. This is primarily due to the inflectional nature of Sanskrit. We find similar trends in the vocabulary of R{\=a}m{\=a}ya\d{n}a\footnote{\url{https://sanskritdocuments.org/sites/valmikiramayan/}} and Digital Corpus of Sanskrit \cite{dcsOliver} as well.


\begin{table*}[t!]
\small
\begin{centering}
\begin{tabular}{ |C {6.9cm}|C {4.9cm}|C{0.7cm}|C{0.5cm}|C{1cm}|}
\hline
  \textbf{Process} & \textbf{Parameters} & \multicolumn{2}{c|}{\textbf{Range}} & \textbf{Step size} \\ \hline
  Gamma Correction (GM) & gamma ($\gamma$)   & 4 & 64 &  4 \\ \hline
  Salt \& Pepper Noise (SPN) (with 50\% salt and 50\% pepper) & percentage of pixels corrupted &0.1\%&1\%& 0.1 \\ \hline
  Gaussian Noise (GN) (mean = $0$)  & standard deviation  & 2.5 & 3.5 & 0.25  \\ \hline
  Erosion (E) (one iteration) & kernel size (m $\times$ m) & 2 & 5 & 1  \\ \hline
  Horizontal perspective distortion (HPD) & image width by image height & 0.3 & 1 & 0.05  \\ \hline
\end{tabular}
\par\end{centering}
\caption{Image pre-processing steps and parameters}
\label{image_params}
\end{table*}

\subsection{Synthetic Generation of training set}
\label{synGen}
Using the text-lines from Bh{\=a}gavatam, we generate synthetic images using ImageMagick\footnote{\url{https://www.imagemagick.org/script/index.php}}. The images were generated with a quality of 60 Dots Per Inch (DPI). The number of pixels along the height for each textline was kept constant at 65 pixels. We add several distortions to the synthetically generated images so as to visually match with the same settings as that of \textit{OCRTest}. Previously, \newcite{Ul-Hasan:2013:WBL:2505377.2505394} used the approach of synthetically generating training data for multilingual OCR solution of theirs.

\begin{figure}[h!]
    \centering
    \includegraphics[width=\linewidth]{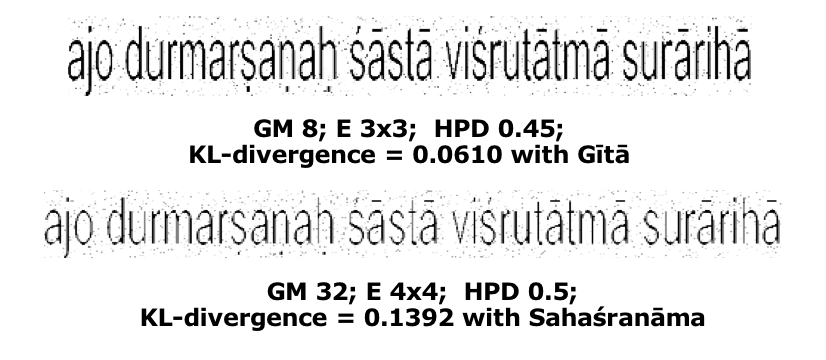}
    \caption{Samples of synthetically generated images. The parameter settings for the distortions are mentioned below the corresponding image.}
    \label{fig:distort}
    \vspace{-1em}
\end{figure}

Table \ref{image_params} shows the different parameters, namely, gamma correction, noise addition, use of structural kernel for erosion and perspective distortion, that we apply sequentially on the images so as to distort and degrade the images \cite{chen2014large}. We use grid search for the parameter estimation for these processes, where those parameters and the range of values experimented with are provided in Table~\ref{image_params}. Finally, we filter 7 (out of 38,400 combinations)  different configurations based on the distribution of Character Recognition Rate (CRR) across the images compared with that of the \textit{OCRTest} using KL-divergence. Among these seven configurations, four are closer to the settings for Bhagavad G{\=\i}t{\=a} and the remaining three for Saha\'{s}ran{\=a}ma. Figure \ref{fig:distort} shows the two different settings (closer to each of the source textbook) for the string \textit{``ajo durmar\d{s}a\d{n}a\d{h} \'{s}\={a}st\={a} vi\'{s}rut\={a}tm\={a} sur\={a}rih\={a}''}, along with their corresponding parameter settings and KL-Divergence. Our training set contains images from all the 7 settings for each of the textline in OCRTrain\footnote{Samples of all the 7 seven configurations are shown in the supplementary material in \S 4}.

\paragraph{Evaluation Metrics}
We use three different metrics  for evaluating all our models. We use Character Recognition Rate (CRR) and Word Recognition Rate (WRR) averaged over each of the sentences in the 430 lines in the test dataset \cite{sankaran2012recognition}. CRR is the fraction of characters recognised correctly against the total number of characters in a line, whereas WRR is the fraction of words correctly recognised against the total number of words in a line. Additionally, we use a sentence level metric, called the acceptability score. The measure indicates the extent to which a sentence is permissible or acceptable to the speakers of the language \cite{lau-clark-lappin:2015:ACL-IJCNLP}. From \newcite{lau-clark-lappin:2015:ACL-IJCNLP}, we use the \textit{NormLP} formulation for the task, as it is found to have a high correlation with the human judgements in evaluating acceptability. NormLP is calculated by obtaining the likelihood of a predicted sentence as per the model, and then normalising it by the likelihood of the string as per a unigram language model trained on a corpus with gold standard sentences. A negative sign is then given to the score. The higher the score, the more acceptable the sentence is.

\subsection{Baselines}

\paragraph{Character Tagger - Sequence Labelling using BiLSTMs} This is a sequence labelling model which uses BiLSTM cells and input is a character sequence \cite{saluja2017error}. We use categorical cross-entropy as the loss function and softmax as the activation function. For dropout, we employ spatial dropout in our architecture. The model consists of 3 layers with each layer having 128 cells. Embeddings of size 100 are randomly initialised and the learnt representations are stored in a character look-up table similar to \newcite{lample-EtAl:2016:N16-1}. In addition to every phoneme in Sanskrit as a class, we add an additional class `no change' which signifies that the character remains as is.  We also experimented with a variant where the final layer is a CRF layer \cite{lafferty2001conditional}. 
We henceforth refer to both the systems as \textit{BiLSTM} and \textit{BiLSTM-CRF}, respectively.

\textbf{Pruned CRFs \cite{schnober2016still}}: They are higher order CRF models \cite{ishikawa2011transformation} that approximate the CRF objective function using coarse-to-fine decoding. \newcite{schnober2016still} adapt the sequence labeling model as a seq2seq model that can handle variable length input-output pairs. 
\newcite{schnober2016still} show that none of the neural seq2seq models considered in their work were able to outperform the Pruned CRF with order-5. The features to the model are consecutive characters within a window of size $w$ in either of the directions of the current position at which a prediction is made. The model is designed to handle 1-to-zero and 1-to-many matches, facilitated by the use of alignment prior to training. We consider all the three settings reported in \newcite{schnober2016still} and report the results for the best setting. The order-5 model which uses 6-grams within a window of 6 performs the best. Henceforth, this model is referred to as \textit{PCRF-seq2seq} (also referred to as PCRF interchangeably).

\textbf{Encoder-Decoder Models}: For the seq2seq model \cite{sutskever2014sequence}, 
we use 3 stacked layers of LSTM each at the encoder and the decoder. Each layer is of 128 dimensions and weighted cross-entropy is used as the loss. We also add residual connections among the layers in a stack \cite{wu2016google}. To further capture the entire input context for making each prediction at the output, we make use of attention \cite{bahdanau2014neural}, specifically Luong's attention mechanism \cite{luong-pham-manning:2015:EMNLP}. We experiment with two variants where \textit{EncDec+Char} uses character level embeddings and \textit{EncDec+BPE} uses embeddings with BPE.

\textbf{CopyNet+BPE}: The model discussed in \S \ref{model}. We use CopyNet+BPE and CopyNet interchangeably throughout the paper.

%% file: 1results.tex
\subsection{Results}

Table \ref{results} shows the results for all the competing systems based on the predictions from \textit{OCRTest}. CopyNet performs the best among the competing systems across all the three metrics and on both the source texts. 
For the G{\=\i}t{\=a} dataset, the models CopyNet and PCRF-Seq2Seq report similar performances. However, Saha\'{s}ran{\=a}ma is a noisier dataset, and we find that CopyNet outperforms all other models by a huge margin. The WRR for the system is double that of the next best system (EncDec) on this dataset. 


\begin{table*}[t!]
\small
\begin{centering}
\begin{tabular}{|c|l|l|c|l|l|c|l|l|c|}
\hline
\textbf{Model}             & \multicolumn{3}{c|}{\textbf{Bhagavad G{\=i}t{\=a}}} & \multicolumn{3}{c|}{\textbf{Saha\'{s}ran{\=a}ma}}  & \multicolumn{3}{c|}{\textbf{Combined}}\\ \hline
\multicolumn{1}{|l|}{}        &  CRR  & WRR & Norm LP  & CRR    &  WRR & Norm LP & CRR & WRR & Norm LP   \\ \hline
  OCR & 84.81\% & 64.40\% & -- & 35.76\% & 0.65\% & -- & 77.88\% & 23.84\% & -- \\ \hline
  BiLSTM & 93.79\% & 68.60\% & -0.553 & 61.31\% & 7.28\% & -1.292 & 85.23\% & 45.60\% & -0.852\\ \hline
  BiLSTM-CRF & 94.68\% & 68.60\% & -0.548 & 65.31\% & 7.28\% & -1.281 & 85.82\% & 45.60\% & -0.847 \\ \hline
  PCRF-seq2seq & \textbf{96.87\%} & 70.56\% & -0.227 & 81.77\% & 9.34\% & -1.216 & 87.94\% & 57.17\% & -0.803 \\ \hline
  EncDec+Char & 91.48\% & 68.00\% & -0.542 & 63.63\% & 15.74\% & -1.321 & 82.51\% & 47.37\% & -0.865 \\ \hline
  EncDec+BPE & 90.92\% & 68.00\% & -0.496 & 61.53\% & 15.74\% & -1.384 & 83.14\% & 45.98\% & -0.842 \\ \hline
  \textbf{CopyNet+BPE} & \textbf{97.01\%} & \textbf{75.21\%} & \textbf{-0.165} & \textbf{87.01\%} & \textbf{33.47\%} & \textbf{-0.856} & \textbf{89.65\%} & \textbf{68.71\%} & \textbf{-0.551}\\ \hline
\end{tabular}
\par\end{centering}
\caption{Performance in terms of CRR, WRR and Norm LP (acceptability) for all the competing models}
\label{results}
\end{table*}

\begin{table*}
\small
\centering

\begin{minipage}{0.33\textwidth}
\centering
   \begin{tabular}{|c|l|l|c|l|l|c|l|l|c|}
\hline
\textbf{} & \multicolumn{1}{c|}{CRR} & \multicolumn{1}{c|}{WRR} \\ \hline
  \textbf{Bhagavad G{\=i}t{\=a}} & 96.80\% & 71.23\% \\ \hline
  \textbf{Saha\'{s}ran{\=a}ma} & 82.81\% & 26.01\%  \\ \hline
  \textbf{Combined} & 87.88\% & 60.91\% \\ \hline
\end{tabular}
\caption{Performance in terms of CRR, WRR for Google OCR}
\label{resultsGoog}
\end{minipage}%
\hfill
\begin{minipage}{0.63\textwidth}
\centering

\begin{tabular}{|c|l|l|l|l|l|l|l|l|c|}
\hline
\textbf{Model}             & \multicolumn{3}{c|}{\textbf{Bhagavad G{\=i}t{\=a}}} & \multicolumn{3}{c|}{\textbf{Saha\'{s}ran{\=a}ma}} & \multicolumn{3}{c|}{\textbf{System errors}} \\ \hline
\multicolumn{1}{|l|}{}        &  Ins  & Del  & Sub    &  Ins & Del & Sub &  Ins & Del & Sub     \\ \hline
OCR                           &  23   & 63   & 1868   &  73 & 696 & 1596 & -- & -- & --     \\ \hline
PCRF                  &  22   & 57   & 641    &  72 & 663 & 932  &  0 & 73 & 209     \\ \hline
\textbf{CopyNet}          &  22   & \textbf{45}   & 629    &  72 & \textbf{576} & 561  &  10 & 5 & 52 \\ \hline
\end{tabular}
\caption{Insertion, Deletion and Substitution errors for OCR, PCRF and CopyNet modes for both the datasets. The system errors are extra errors added by the respective systems.}
\label{tab:OCR_err}

\end{minipage}
\end{table*}

\begin{figure}[!htb]
\small
\centering
\begin{subfigure}[b]{\linewidth}
    \centering
	\includegraphics[width=\linewidth]{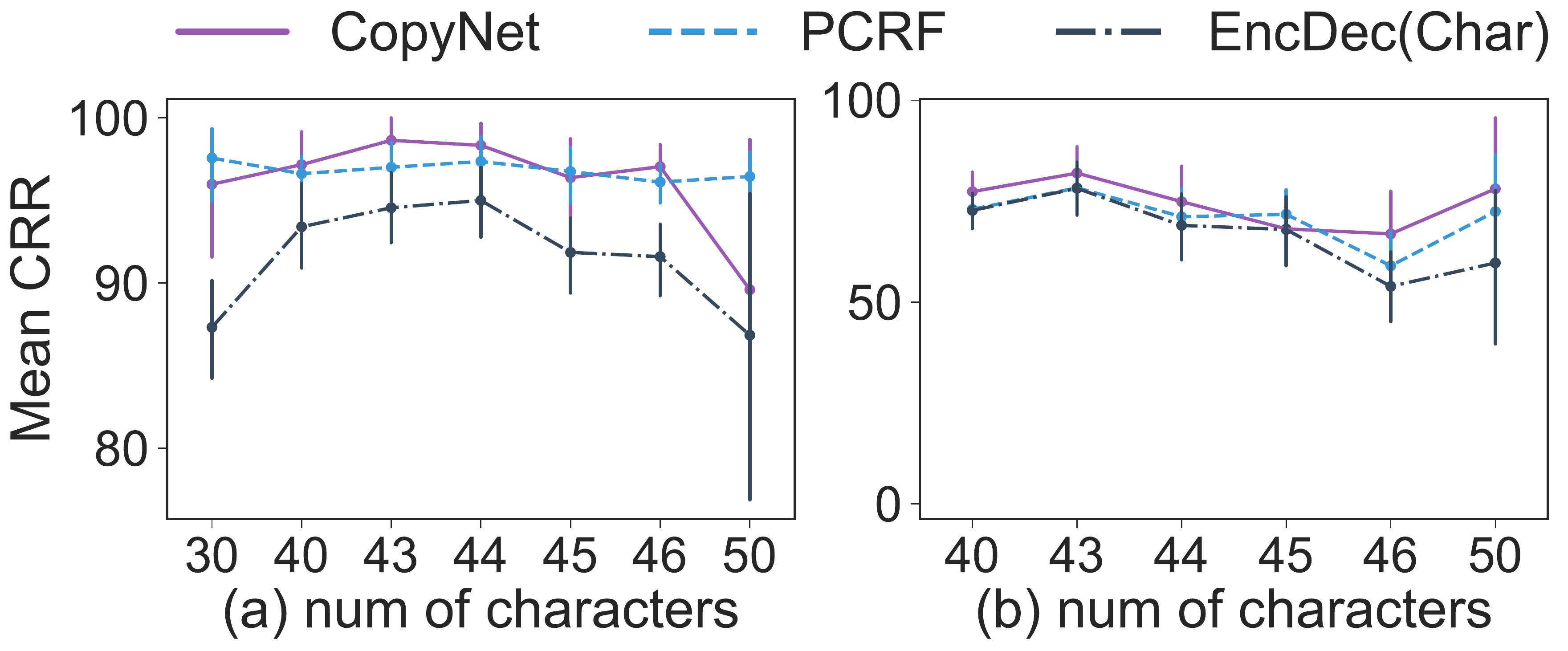}
\end{subfigure}
\begin{subfigure}[b]{\linewidth}
    \centering
  	\includegraphics[width=\linewidth]{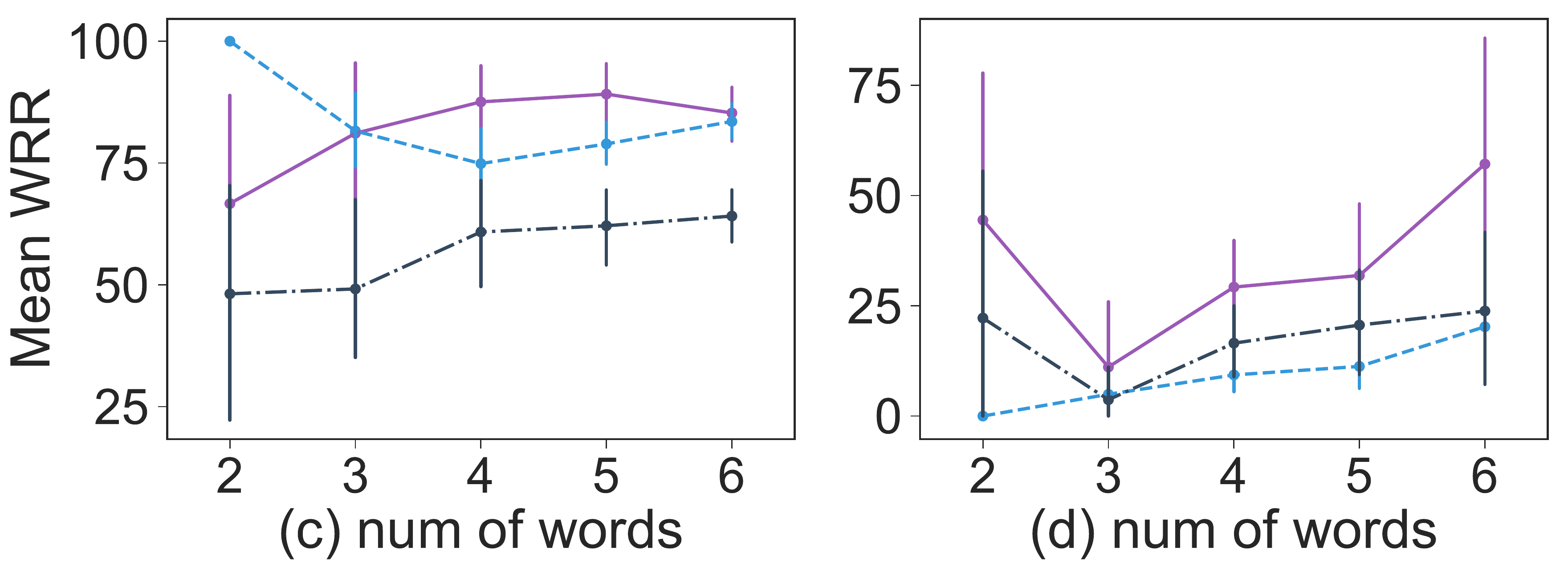}
\end{subfigure}
\caption{(a) and (b) show CRR for G{\=\i}t{\=a} and Saha\'{s}ran{\=a}ma respectively, for the competing systems. (c) and (d) shows WRR for G{\=\i}t{\=a} and Saha\'{s}ran{\=a}ma, respectively. All the entries with insufficient data-points were merged to the nearest smaller number.}
\label{fig:crr-wrr}
\end{figure}

\paragraph{System performances for various input lengths:}
From Figure \ref{fig:crr-wrr}a, it can be observed that the performance in terms of CRR for CopyNet and PCRF is robust across all the lengths on strings from G{\=\i}t{\=a} and never goes below 90\%. 
For Saha\'{s}ran{\=a}ma, as shown in Figure \ref{fig:crr-wrr}b, CopyNet outperforms PCRF across inputs of all the lengths except for one setting. But, in the case of WRR, CopyNet is the best performing model across all the lengths as shown in Figure \ref{fig:crr-wrr}d.  




\paragraph{Error type analysis} 
In Table \ref{tab:OCR_err}, we analyse the reduction in specific error types for PCRF and CopyNet after the alignment of the predicted string with that of the ground truth in terms of insertion, deletion and substitution. We also report the system induced errors, where a correct component at the input (OCR output) is mispredicted to a wrong output by the model. CopyNet outperforms PCRF in correcting the errors and it also introduces lesser number of errors of its own. Both CopyNet and PCRF \cite{schnober2016still} are seq2seq models and can handle varying length input and output. Both the systems perform well in handling substitution errors, the type which dominated the strings in \textit{OCRTest}, though neither of the systems was able to correct the insertion errors. Insertion can be seen as a special case of 1-to-many insertion matches, which both systems are ideally capable of handling. We see that for Saha\'{s}ran{\=a}ma, CopyNet corrects about 17.24 \% of the deletion errors as against $<$5\% of the deletion errors corrected by PCRF. 

\begin{figure}[!htb]
\small
\begin{subfigure}[b]{0.9\linewidth}
    \centering
  	\includegraphics[trim={0 0cm 0cm 0cm},clip,width=\linewidth]{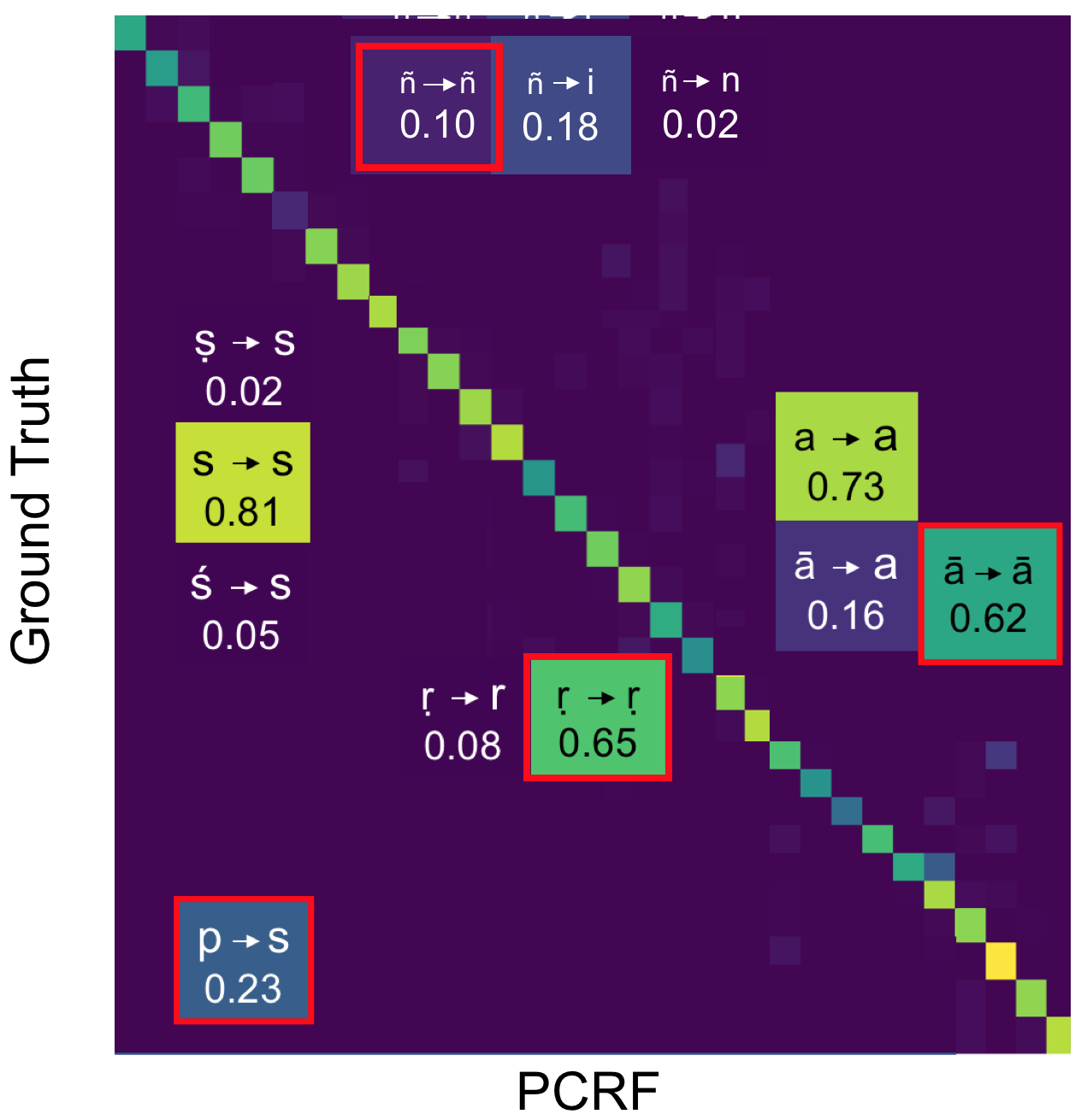}
    \caption{}
  	\label{pcrf}
\end{subfigure}
\begin{subfigure}[b]{0.9\linewidth}
    \centering
	\includegraphics[trim={0 0cm 0cm 0cm},clip,width=1\linewidth]{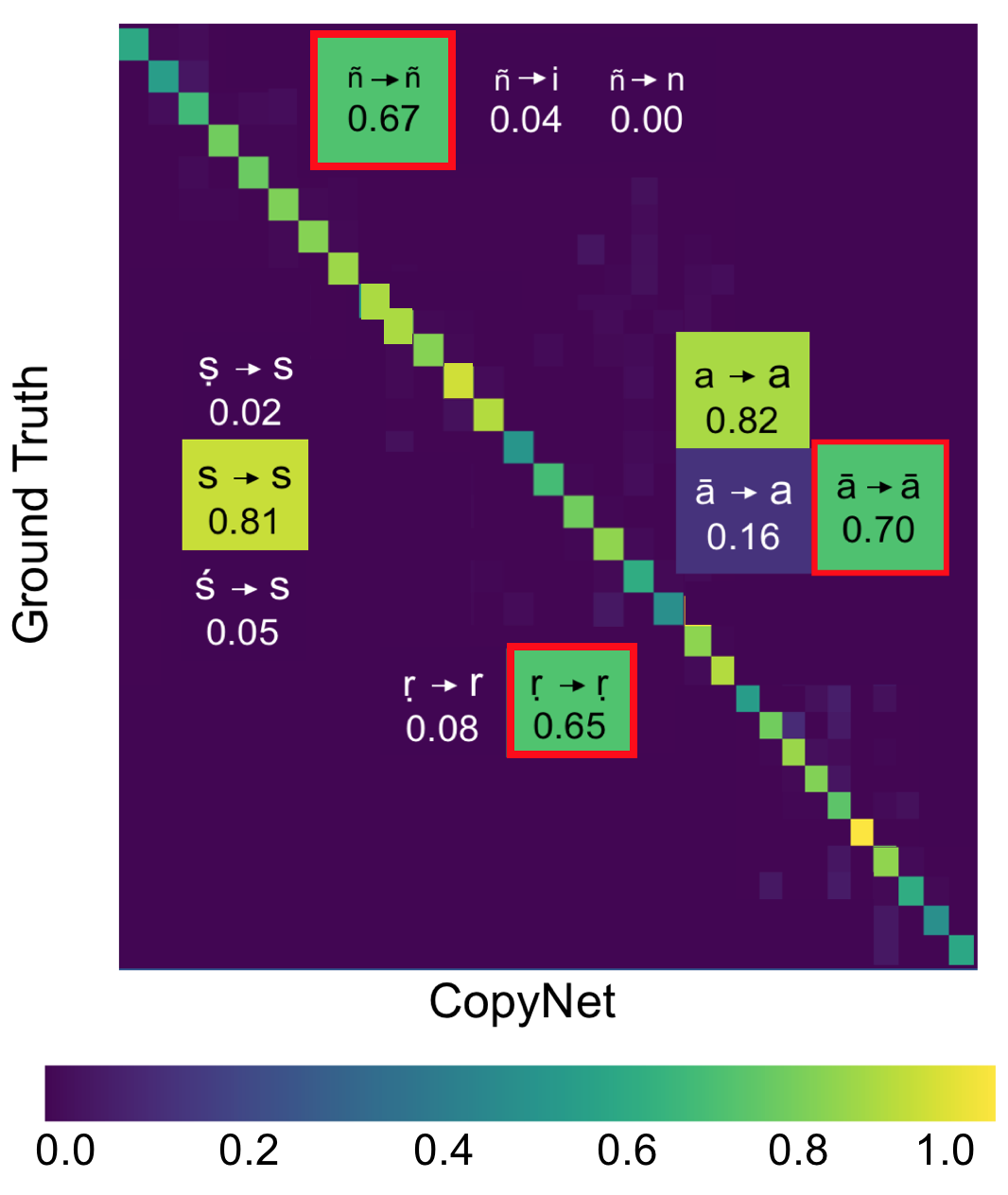}
    \caption{}
  	\label{copy}
\end{subfigure}
\caption{Heatmap for occurrences of majorly confusing character pairs between Ground Truth and predictions of (a) PCRF model (b) CopyNet model}
\label{subHeat}
\vspace{-1.2em}
\end{figure}


Since there exist 14 graphemes in Sanskrit alphabet which are not present in the English alphabet, all 14 of them get substituted to a different grapheme by the OCR. While most of them get substituted to an orthographically similar character such as \={a} $\rightarrow$ a and \d{h} $\rightarrow$ h, we find that \~{n} $\rightarrow$ i does not fit the scheme, as shown in Figure \ref{fig:ocr_errors}. In the majority of the cases, CopyNet predicts them to the correct grapheme. But PCRF still fails to correct the OCR induced confusion for \~{n} $\rightarrow$ i in the majority of the instances. Additionally, we find that PCRF introduces its own errors, for instance it often mispredicts p $\rightarrow$ s. Figure \ref{subHeat} shows the overall variations in both the systems as compared to  Figure \ref{fig:ocr_errors} for OCR induced errors.

\begin{figure}[!htb]
\centering
  	\includegraphics[trim={0 0cm 0cm 0},clip,width=\linewidth]{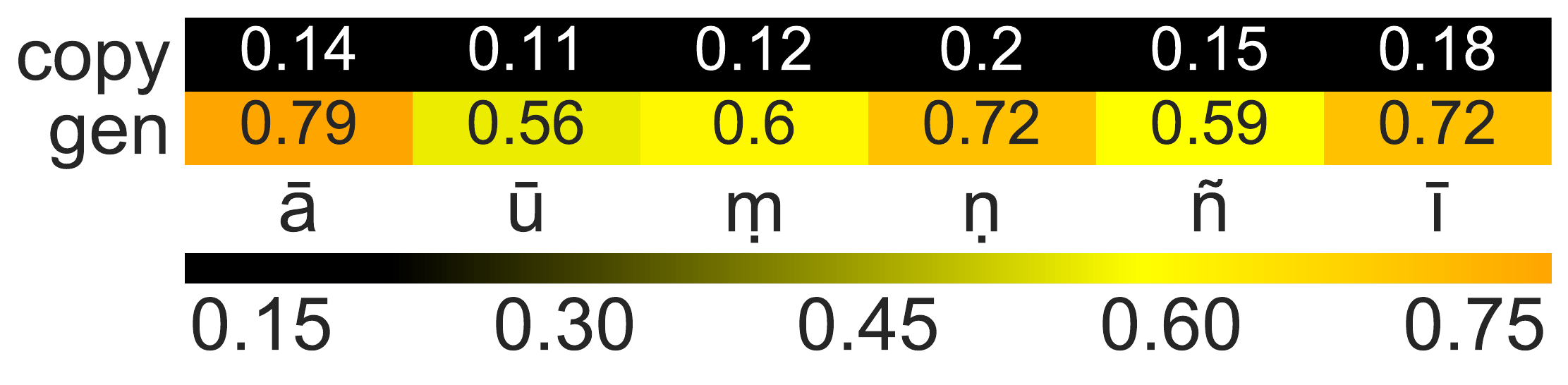}
\caption{Heatmap of mean copy score (copy) and mean generate score (gen), respectively for 6 (of 14) graphemes not present in the English alphabet.}
\label{fig:copygen_global}
\vspace{-1em}
\end{figure}


\begin{figure}[!htb]
\centering
  	\includegraphics[trim={0 0cm 0cm 0},clip,width=\linewidth]{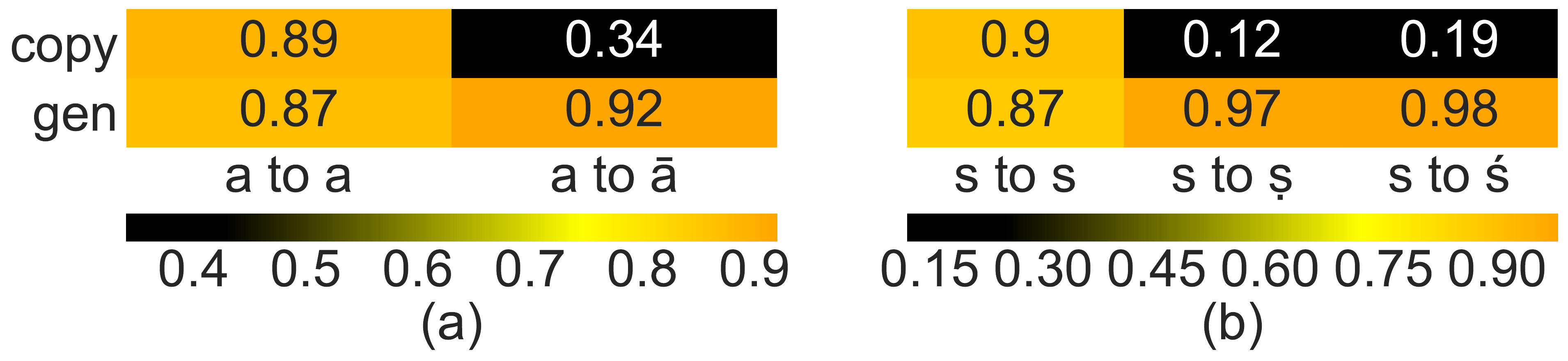}
  	\label{copy_gen_a_s}
  	\vspace{-5mm}
\caption{mean copy and generate scores for different predictions from (a) `a' and (b) `s'.}
\label{copygenHeat}
\vspace{-1em}
\end{figure}

\paragraph{Copy or generate?} For the 14 graphemes, missing at the encoder (input) but present at the decoder side during training, those predictions have to happen with high values of generate probability in general. We find that not only the average generate probability for such instances is high but also the copy probability is extremely low. For the remaining cases, we find that both generate and copy probability are higher. But it needs to be noted that the prediction is made generally by summing of both the distributions and the distributions are not complementary to each other. A similar trend can be observed in Figure \ref{fig:copygen_global} as well. For example in the case of a $\rightarrow$ \={a}, only the generate probability is high. But, for a $\rightarrow$ {a}, both the copy and generate probability scores are high.   

\paragraph{Effect of BPE and alphabet in the vocabulary} We further investigate the effect of our vocabulary which is the union of the alphabet in Romanised Sanskrit and what is learnt using BPE. We train the model with only the alphabet as vocabulary and find the CRR and WRR for the combined test sentences to be 86.1\% and 66.09\%, respectively. When using the original BPE vocabulary, we find that there is a slight increase in the performance than the current vocabulary with a CRR and WRR of 89.53\% and 68.11\%, respectively\footnote{Please refer to \S 5 of the Supplementary material for the performance table}. We also find that the current setting performs better than a model that takes word level input. The word level model shows a drop in the performance with a CRR and WRR of 86.42\% and 66.54\%, respectively.

\paragraph{Performance comparison to Google OCR:} Google OCR is probably the only available OCR that can handle Romanised Sanskrit. We could not find the architecture of the OCR or whether the service employs post-OCR text correction. We empirically compare the performance of Google OCR on \textit{OCRTest} with our model. Table \ref{resultsGoog} shows the results for Google OCR. Overall we find that CopyNet outperforms Google OCR across all the metrics. We find that Google OCR reports a similar CRR for G{\=\i}t{\=a} with that of ours, but still reports a lower WRR than ours. The system performs better than PCRF in all the metrics other than CRR for G{\=\i}t{\=a}.

\paragraph{Image quality:} Our training set was generated with a quality of 60 DPI for the images. We generate images corresponding to strings in \textit{OCRTrain} with DPI of 50 to 300 in step sizes of 50 for a sample of 500 images. We use noise settings as shown in Figure \ref{fig:distort}. The OCR output of the said strings remained as is with that of the one generated with a DPI of 60. This experiment can be seen as a proxy in evaluating the robustness of the OCR to various scanning qualities of input. Our choice of DPI as 60 was based on the lowest setting we observed in digitisation attempts in Sanskrit texts. 

\paragraph{Effect of adding distortions to the synthetically generated images:} Table \ref{results} shows the system performance after training our model on data generated as per the procedure mentioned in Section \ref{synGen}. Here, we make an implicit assumption that we can have access to a sample of textline images annotated with the corresponding text from the manuscript for which the Post-OCR text correction needs to be performed. This also mandates retraining the model for every new manuscript. We attempted for a more generalised version of our model, by using training data where the image generation settings are not inspired from the target manuscript for which the task needs to be performed.  Using the settings from \cite{chen2014large} for inducing noise, we generated 10 random noise configurations. Here the step sizes were fixed at values such that each parameter, except erosion (E), can assume 5 values each uniformly spread across the corresponding ranges considered. From a total of 2500 (5$\times$5$\times$5$\times$5$\times4$) configuration options, 10 random settings were chosen. Every textline was generated with each of the 10 different settings. The resulting model using CopyNet produced a CRR of 89.02\% (96.99\% for G{\=i}t{\=a} and 85.62\% for Saha\'{s}ran{\=a}ma) on the test set, which is close to the reported CRR of 89.65 in Table \ref{results}. The noise ranges chosen are used directly from \cite{chen2014large} and are not influenced by the test data in hand. 

We also experimented with a setting where no noise was added to the synthetically generated images and the images were fed to the OCR. We obtained a CRR of 80.12\% from OCR, where the errors arose mostly from the missing graphemes in the alphabet getting mispredicted to a different grapheme. CopyNet after training with the text so generated reported a CRR of 86.81\% (96.01\% for G{\=i}t{\=a}, 75.78\% for Saha\'{s}ran{\=a}ma) on the test data. 
   

\paragraph{Human judgement survey:} In this survey\footnote{More details at \S 6 of Supplementary material}, we evaluate how often a human can recognise the correct construction by viewing only the prediction from one of the systems. We also evaluate how fast a human can correct them. We selected 15 constructions from Saha\'{s}ran{\=a}ma, and obtained the system outputs from the OCR, CopyNet and PCRF for each of these. The average length of a sentence is 41.73 characters, all ranging between 23 and 47 characters. A respondent is shown a system prediction (system identity anonymised) and is asked to type the corrected string without referring to any sources. A respondent gets 15 different strings altogether, 5 each from each of the three systems. We consider responses from 9 participants where all of them at least have an undergraduate degree in Sanskrit linguistics. Altogether from 3 sets of questionnaires, we have 45 strings (3 outputs for a given string). Every string obtained 3 impressions. We find that a participant on an average could identify 4.44 sentences out of 5 from the CopyNet, while it was only 3.56 for PCRF and 3.11 for the OCR output. The average time taken to complete the correction of a string was 81.4 seconds, 106.6 seconds and 127.6 seconds for CopyNet, PCRF and OCR, respectively.

\section{Conclusion}
In this work, we proposed an OCR based solution for digitising Romanised Sanskrit. Our work acts as a Post-OCR text correction approach and is devoid of any OCR-specific feature engineering. 
We find that the use of copying mechanism in encoder-decoder performs significantly better than other seq2seq models for the task. Our model outperforms the commercially available Google OCR on the Saha\'{s}ran{\=a}ma texts. From our experiments, we find that CopyNet performs stably even for OCR outputs with a CRR as low as 36\%. 
Our immediate research direction will be to rectify insertion errors which currently are not properly handled. Also, there are 135 languages which directly share the Roman alphabet but only 35 of them have OCR system available. Our approach can be easily extended to provide a post-processed OCR for those languages.

\section*{Acknowledgements}
We are grateful to Amba Kulkarni, Arnab Bhattacharya, Ganesh Ramakrishnan, Rohit Saluja, Devaraj Adiga and Hrishikesh Terdalkar  for helpful comments and discussions related to Indic OCRs. We would like to thank Madhusoodan Pai, Sanjeev Panchal, Ganesh Iyer and his students for helping us with the human judgement survey. We thank the anonymous reviewers for their constructive and helpful comments, which greatly improved the paper.